# Evolving Ensemble Fuzzy Classifier

Mahardhika Pratama, *Member, IEEE*, Witold Pedrycz, *Fellow, IEEE*, Edwin Lughofer

***Abstract*— the concept of ensemble learning offers a promising avenue in learning from data streams under complex environments because it better addresses the bias and variance dilemma than its single-model counterpart and features a reconfigurable structure, which is well-suited to the given context. While various extensions of ensemble learning for mining non-stationary data streams can be found in the literature, most of them are crafted under static base-classifier and revisit preceding samples in the sliding window for a retraining step. This feature causes computationally prohibitive complexity and is not flexible enough to cope with rapidly changing environments. Their complexities are often demanding because they involve a large collection of offline classifiers due to the absence of structural complexities reduction mechanisms and lack of an online feature selection mechanism. A novel evolving ensemble classifier, namely Parsimonious Ensemble (pENsemble), is proposed in this paper. pENsemble differs from existing architectures in the fact that it is built upon an evolving classifier from data streams, termed Parsimonious Classifier (pClass). pENsemble is equipped by an ensemble pruning mechanism, which estimates a localized generalization error of a base-classifier. A dynamic online feature selection scenario is integrated into the pENsemble. This method allows for dynamic selection and deselection of input features on the fly. pENsemble adopts a dynamic ensemble structure to output a final classification decision where it features a novel drift detection scenario to grow the ensemble's structure. The efficacy of the pENsemble has been numerically demonstrated through rigorous numerical studies with dynamic and evolving data streams where it delivers the most encouraging performance in attaining a tradeoff between accuracy and complexity.**

## I. INTRODUCTION

THE data-intensive era where data are collected continuously at a fast rate under dynamic and evolving environments opens a new promising research direction to process data streams efficiently [1], [2]. Unlike a classical paradigm in machine learning where a dataset is utilised to construct hypothesis and is executed over multiple passes, data streams requires a strictly online learning framework with a low memory requirement and even if possible with no memory at all – one-pass learning mode. Another challenging trait of data streams lies in their non-stationary characteristics [3] where the data do not follow static and predictable distributions and contains a variety of concept drifts [4], [5]. These facts make a retraining phase when incorporating a new sample to an old dataset impossible to be performed because it leads to the so-called catastrophic forgetting [6] of previously valid knowledge and is not scalable when dealing with massive data streams. For data stream analytics, a learning machine should satisfy the following requirements [18]:

1. Data arrive one by one or chunk by chunk;
2. Data are learned in a single-scan without revisiting previously learned samples;
3. The total number of samples is unknown before process runs;
4. A learning machine must have drift handling capability regardless of how slow, rapid, abrupt, gradual, temporal, local or global change in data stream is;

Evolving Intelligent System (EIS) provides a unique solution for data stream mining because a strictly one-pass learning procedure involved here has delivered great success to cope with time-critical applications where data streams are generated at a very fast sampling rate [7]. Furthermore, EIS adopts an open structure where its components can be automatically generated, pruned, merged and recalled on the fly [8], [9] and can be well-suited to a given problem. This trait reflects the true data distributions and tracks changing data distributions [10]. EIS has transformed to be one of the most active research area in the computational intelligence research as evidenced by the number of published works in this area [71]. Nonetheless, EIS is typically built upon a single classifier architecture which often does not produce adequate accuracy for complex problems [11], [35]. In fact, from classical batch learning perspective, it is well-known that ensemble classifiers outperform single base classifiers in case of high noise levels and a low number of available training samples [12] because they can better resolve the bias-variance dilemma due to proper subspace and data exploration using weak classifiers [13]. While few works about a synergy between EIS and an ensemble structure can be found in the literature [14], [15], most of them utilise a static ensemble architecture, which should be predetermined in advance. Although diversity of base classifiers can be guaranteed by varying user-defined parameters or applying different data partitions to base classifiers, the issue of concept drifts remains an open challenge because of their fixed structure.

This paper was submitted on May, 21st, 2017. The first author acknowledges the support of NTU start-up grant and MOE Tier 1 research grant. The second author acknowledges the support of the Natural Sciences and Engineering Research Council of Canada (NSERC) and Canada Research Chair (CRC) in Computational Intelligence. The third author acknowledges the support of the Austrian COMET-K2 programme of the Linz Center of Mechatronics (LCM), funded by the Austrian federal government and the federal state of Upper Austria.

M. Pratama is with School of Computer Science and Engineering, Nanyang Technological University, Singapore (e-mail: pratama@ieee.org).

W. Pedrycs is with Department of Electrical and Computer Engineering, University of Alberta, Canada, Department of Electrical and Computer Engineering, Faculty of Engineering, King Abdul-Aziz University, Saudi Arabia, Systems Research Institute, Polish Academy of Science, Warsaw, Poland (e-mail: wpedrycz@ualberta.ca).

E. Lughofer is with Department of Knowledge-Based Mathematical Systems, Johannes Kepler University, Linz, Austria, (e-mail: edwin.lughofer@jku.at).

The ensemble learning concept uses combination of individual base classifiers with a modularity principle, where it enables a dynamic evolution of the ensemble structure [12]-[19]. The key of ensemble learning lies in the diversity of base classifiers, which makes them more robust to various forms of uncertainty in data streams (such as significant noise levels). Nonetheless, one must bear in mind that the diversity of an ensemble classifier might be counterproductive in realm of data streams because it opens the door for outdated base-classifiers in the ensemble structure. Adaptability of the ensemble classifier plays a vital role to the success of ensemble learning because it formulates mechanisms how an ensemble classifier adapts itself when changing data distributions are presented [18]. The ensemble classifier can also be distinguished into two groups: active and passive approach. The passive approach relies on continuous updates of its components and assumes that the concept drifts occur in the ongoing fashion; the active approach is equipped by a dedicated drift detection mechanism in which it is restructured and parameters are fine-tuned when a drift is captured [19]. In practise, the drift detection mechanism plays key to role to alert operators for possible changing system behaviours and to identify whether a change causes catastrophic effect to operation's cycle – vital for process's safety.

To the best of our knowledge, **local concept drift, curse of dimensionality, and structural complexity** are the three open issues in the current literatures. In case of local concept drift, changes do not ensue in the whole feature space rather in some local regions only with different rates and severities [20] [21]. It remains an open question because existing ensemble classifiers are mostly constructed using a batch classifier or accumulate already seen samples in the sliding window for retraining steps and considers only the global change in data distribution. Although ensemble algorithms like DELA [16] is excluded from the local concept drift bottleneck due to its three levels of adaptivity, namely structural adaptivity, combination adaptivity, model adaptivity, it suffers from the absence of a dedicated drift detection method [16]. Furthermore, the structural complexities of existing ensemble classifiers are considerable because they usually involve a large number of base classifiers to assure acceptable accuracy. Most of them suffer from the absence of a structural complexity reduction mechanism which alleviates complexities of ensemble classifiers [22]. Existing ensemble classifiers also assume that input features are pre-selected in the pre-processing steps. This issue hinders its viability in the time-critical applications where data streams are generated continuously in a fast sampling rate which makes an iterative pre-processing step impractical. Furthermore, pre-recorded data are often irrelevant in the later stage because of rapidly changing environments.

A novel ensemble learning algorithm, namely Parsimonious Ensemble (pENsemble), is proposed in this paper. pENsemble features an open structure where a local expert is created and pruned dynamically under strictly one-pass learning mode. It is constructed with a recently published evolving classifier, namely *Parsimonious Classifier (pClass)* [24]. An evolving classifier strengthens the adaptive nature of evolving ensemble because it handles a local concept drift better than a classical batch classifier with its dynamic and online paradigm. It features an open structure paradigm which is self-evolving to track variations in the local data space. The original pClass is, however, created using a generalized TSK fuzzy rule imposing considerable computational and space complexities if used under the roof of the ensemble classifier. pClass is here implemented using both standard axis-parallel Gaussian fuzzy rule and the multivariate Gaussian function. Because pClass is originally designed with the multivariate Gaussian function, pClass with the axis-parallel ellipsoidal rule can be realized with ease by putting forward diagonal covariance matrix instead of non-diagonal one. pENsemble works fully in the single-pass learning mode, which is well-suited to the online life-long learning scenario. pENsemble is also equipped with a dynamic feature selection scenario which can address a high input dimensionality and to the best of our knowledge is absent from the majority of existing ensemble classifiers. The final class prediction of pENsemble is inferred by a dynamic ensemble paradigm [25] which dynamically grow, shrink and adjust the weights of local experts to data streams. The dynamic ensemble concept is inspired by the evolving trait of DWM [34] but different criteria are applied to perform the structural learning scenarios of pENsemble. pENsemble puts forward three new learning components as follows:

- *Online Drift Detection Scenario*: pENsemble adopts a dynamic ensemble structure where a new local expert can be added when a concept change presents in the data streams [26]. This procedure is governed by a non-parametric drift detection method derived from the concept of Hoeffding's bounds [27]. This method monitors the performance metric and sends a warning signal when a significant variation is identified. This method is threshold-free and relies on some probability inequalities under assumption of independent, univariate and bounded random variables which has been theoretically proven. This learning feature lowers the ensemble complexity because the ensemble size expands on demands only and is independent from the number of data streams.
- *Ensemble Pruning Scenario*: pENsemble presents an ensemble pruning scenario which is crafted from the notion of localized generalization error [28]. This method estimates generalization performance of a local expert [29] and determines local experts to be pruned [30]. This technique analyses the upper bound of error of a local expert within Q neighbourhood which reflects the generalization power of a local expert. This notion is proposed in [28]-[31] under a radial basis function neural network (RBFNN) and is adapted to the working principle of pENsemble here applying a TSK neuro fuzzy local expert, namely pClass.
- *Online Feature Selection Scenario*: pENsemble is capable of performing an online feature selection scenario using the so-called Generalized Online Feature Selection (GOFS) method, an extension of the OFS method in [32]. The advantage of GOFS over its counterparts lies in its capability for selection and deselection of input attributes on the fly by assigning crisp values (0 or 1). This allows flexibility in the feature selection process and avoids the discontinuity bottleneck because an input variable can be recovered again in the future when needed [33]. Another salient feature of the GOFS concept is seen in its aptitude in handling partial input information which relieves computational and storage burdens because a learning process does not necessarily start from a full-scale input variables.

This paper conveys the following four major contributions as follows: 1) a novel ensemble learning algorithm inspired by a seminal work, namely DWM [34], is proposed. It modifies DWM with the introduction of a drift detection scenario, an ensemble pruning scenario, an online feature selection and an evolving local expert; 2) pENsemble puts forward a new perspective of a fully evolving ensemble learning concept where it is evolving in both ensemble level and base-classifier level; 3) three novel learning modules, namely the drift detection method, the ensemble pruning scenario, and the online feature selection, are proposed; 4) the efficacy of pENsemble was numerically validated using 15 real-world and synthetic data streams. It was compared with the state-of-the art classifiers showing that pENsemble outperformed its counterparts in terms of accuracy and complexity.

The paper is structured as follows: Section 2 outlines literature survey over current ensemble learning algorithms and evolving learning algorithms, Section 3 discusses architecture and learning policy of pENsemble, Section 4 elaborates on the working principles of the base classifier – pClass, Section 5 describes numerical studies and comparisons with prominent algorithms, concluding remarks are drawn in the last section.

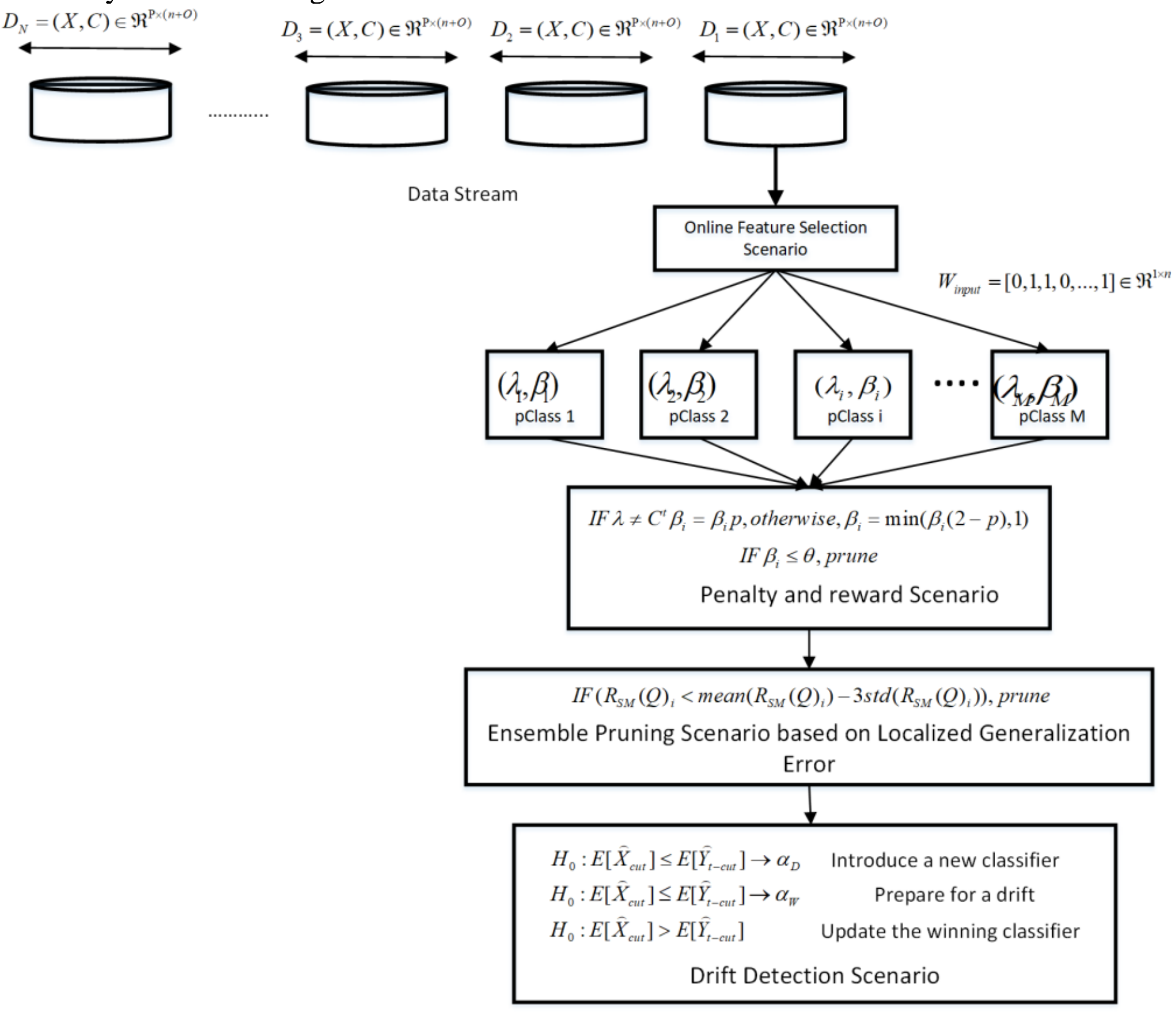

Fig 1. Architecture of pENsemble

## II. RELATED WORKS

Research in the area of EIS has started with algorithmic development of a number of works. Evolving rule-based model exemplifies the EIS concept using the incremental unsupervised learning [37]. DENFIS in [9] is another early example of EIS which combines the working principle of TSK fuzzy system and the Evolving Clustering Method (ECM). Angelov and Filev proposed the so-called eTS [7] which benefits from the data potential theory forming an evolving version of the mountain clustering. This work is modified for a classification problem [65], [66] and has formed the first evolving classifier, termed eClass. The term EIS has not been however formalised until the clarification in [71] since the term "evolving" is sometime confused with the well-known term of evolutionary computation. Motivated by significant progress in real-time data collection and capture, the notion of EIS has gained popularity in the community because it has been shown effective in addressing lifelong learning situation and non-stationary environments. Several extensions and variations of EIS have been put forward in the literature [39], [40], [67]-[70]. An evolving version of Vector quantization was designed in [41] and is algorithmic backbone of FLEXFIS [42], which was later extended to a more robust version including rule merging in [43], generalized rules and an incremental feature weighting mechanism in [44] – an extension of input pruning scenario in [72]. A generalized TSK fuzzy rule was put forward in [45]-[47] and generates a non-axis parallel ellipsoidal cluster, which happens to have better coverage and flexibility than conventional fuzzy rules [44]. Pratama et al in [47] developed the theory of rule statistical contribution borrowing the concept of hidden neuron statistical contribution in [48], [49]. EIS has also been implemented for control system. In [78], adaptive fuzzy tracking control for MIMO uncertain system with guaranteed closed-loop stability was proposed. This approach is extended to take into account various uncertainties [79].

Evolving Ensemble (eEnsemble) was proposed in [14] where it makes use of eTS [7] as a base-classifier and is realised under different configurations of the ensemble classifier. This work was extended in [50] where eStacking is put forward using the concept of stacking ensemble. A parallel implementation of TEDAClass was proposed [69]. This work can be classified as an ensemble in a strict sense where data are distributed in a number of computing nodes. In [70], an ensemble of deep learning classifiers was designed for handwriting recognition and adopted the concept of data parallerization as with [69]. The all-pair classifier in [50] can be also grouped as an ensemble approach. It is solely concentrated on a class decomposition approach for multi-class problems in order to reduce class imbalance. Notwithstanding that the EIS has been well-established in the literature, it still deserves in-depth investigation because of at least three reasons 1) vast majority of EIS is constructed in the single model framework having low diversity. The ensemble learning concept is well-known for its powerful generalization power because it address the bias-and-variance better and produces a model with high diversity covering a rich data region; 2) The use of evolving base classifier in the ensemble structure has been initiated in [14], [15], [50], [69], [70] but it relies on a static ensemble structure which is predetermined during the training process; 3) Existing EISs are categorized as a passive approach in handling concept drift because changing data distributions are overcome by continuously adapting a classifier. It lacks of capability to signal the presence of concept drift and to identify the type of drift. Such trait plays vital role in practice because it provides a feedback to an operator whether a drift is alarming or not.

## III. Learning Policy of pENsemble

This section concerns the learning scenarios of pENsemble including ensemble structure, learning procedure, and complexity analysis. Overview of pENsemble learning scenarios is depicted in Fig. 1.

### A. *Ensemble Structure*

pENsemble is developed under a generalized working framework of the DWM in which its working principle is displayed in Algorithm 1. pENsemble stores a collection of local experts, which can be automatically generated when a drift is detected and pruned when it is no longer relevant to capture current data trends [34]. An evolving algorithm, namely pClass, is deployed as a base learner which implements an open structure paradigm and is created under the MIMO architecture [24]. That is, each rule possesses multiple consequents representing each class and the final output is inferred from that generating the maximum output. The reason behind the choice of the MIMO architecture is its aptitude in handling the class overlapping because each class is looked after by a unique rule consequent. Each local expert is assigned with a voting weight $w_i$ dynamically adjusted by a decreasing factor $p_i$ which penalises a local expert when an incorrect prediction is made. A local expert is pruned if its weight falls below a certain threshold $\beta_1$. Despite the fact that the penalty scenario is necessary to keep the ensemble structure relevant to up-to-date context, it compromises diversity of ensemble. To correct this shortcoming, the weight of a local expert is augmented when it makes correct prediction to maintain the ensemble's diversity and to open possibility for a local expert to pick up again - such mechanism plays vital role when dealing with cyclic drift. In addition, pENsemble is equipped with another rule pruning scenario which measures the generalization potential of a local expert based on a localized generalization error principle.

**Algorithm 1:** Parsimonious Ensemble

*Input*

$D=(X\in\Re^{\mathrm{P}\times n},C^{\mathrm{P}\times O}),n,O,\mathrm{P}$ are a pair of data chunk, the number of input dimension and the number of output dimension, and a data chunk size

$p,y_i,\beta_i$ are a decreasing factor, an *i-th* local expert, a weight of *i-th* local expert

$\widehat{C},\lambda,\sigma\in\Re^{1\times O},\theta$ are global and local predictions, sum of weighted predictions for each class, and pruning threshold

a data chunk $D\in\Re^{\mathrm{P}\times(n+O)}$ is received

*Output*

$\widehat{C}$ global prediction

**For** $t=1,\ldots,\mathrm{P}$ // loops over all examples in the data chunk

Execute the feature selection mechanism to sample the *B* most relevant samples. This scenario aims to address a high input dimensionality – Section 3.B.3

**IF** the ensemble network is empty

$M=1$ // create the first local expert

$\beta_1=1$ // initialize the weight of a local expert

**End**

$\sigma=0$

**For** $i=1,\ldots,M$ // loop over local experts

$\lambda=\max_{j=1,\ldots,O}(y_{i,j})$ // elicits the local prediction

**IF** ( $\lambda\neq C^t$ )

$y_i=\beta_i y_i$ // decreases the weight of a local expert when it predicts incorrectly

$\beta_i=\beta_i p$

**Else**

$\beta_i=\min(\beta_i(2-p),1)$

**End**

$\sigma_\lambda=\sigma_\lambda+y_i$

**End**

$\widehat{C}=\max_{\lambda=1,\ldots,O}(\sigma_\lambda)$ // Produces the global prediction

$\beta_i=\frac{\beta_i}{\sum_{i=1}^{M}\beta_i}$ // normalises the weight

**IF** $\beta_i<\theta$

Prune *i-th* local expert // Prune the local expert with a low weight

**End**

**For** $i=1,\ldots,M$

Calculate the localized generalization error (5) to estimate generalization power of a local expert. A local expert with poor generalization capability is removed - Section 3.B.2

**IF** (7)

Discard *i-th* local expert

**End**

**End**

Undertakes the drift detection method to determine suitable learning actions whether a new classifier should be introduced, a learning process is committed to update the winning classifier, or no learning process is carried out – Section 3.B.1.

pENsemble starts its learning scenario from scratch with no base classifier at all. The first base classifier is initialized using the first data chunk. The ensemble structure grows automatically when changing data distributions are seen. The performance of individual local experts are assessed and a penalty is imposed using the decreasing factor $p_i$ when misclassification is made by using a local expert whereas a reward is granted by increasing its voting weight when correct prediction is returned. After carrying out this procedure, the online concept drift detection method is performed. The drift detection strategy relies on the concept of Hoeffding's bounds to determine the drift's level [27]. The statistical process control approach is integrated to monitor dynamic of data streams [53] and classifies system behaviours into three stages, namely normal, warning and drift. A new base classifier is created using new data streams only when a drift level is reached. A weight of a new learner is initialized to 1. The final output of an ensemble classifier is inferred from a class having the highest accumulated weight. The output of each local learner is weighted by its corresponding weight. All outputs are combined to arrive at a weighted sum of each class. The weight of base classifiers are normalized to assure the partition of unity and the normalization step aims to avoid a new classifier to outweigh old classifiers. Note that pENsemble still aligns to the one-pass learning concept because it learns a data-chunk in a single scan without revisiting previous data chunks and without an iterative learning of a data chunk.

*B. Learning Algorithm of Parsimonious Ensemble*

This section focusses on learning procedure of pENsemble which encompasses the drift detection strategy, the ensemble pruning strategy and the online feature selection strategy.

*1) Drift Detection Method*: The drift detection scenario is vital in the pENsemble because it controls the ensemble complexity. It allows an ensemble structure to expand its size when an uncharted training region comes into picture [19]. An online non-parametric drift detection method is integrated using the Hoeffding's inequalities to determine acceptable level of concept changes in data streams [27]. This method is capable of capturing significant distributional changes in data streams in the one-pass mode and is confirmed by solid theoretical guarantees in [27]. It does not rely on any assumption of probability density function rather the performance metrics is regarded as independent and bounded random variables. It is worth mentioning that the drift handling strategy in [23] does not specifically detect the exact time period where a drift presents since it is derived from the forgetting concept – categorized as a passive approach.

The drift detection scenario starts by monitoring statistics of data streams and defines three conditions: stable – there seems to be no change, warning – a possible concept drift may appear and drift – the drift is clearly identified. The underlying task of the drift detection method is to not only pinpoint when the drift occurs in data streams but also to track the transition between stable condition to drift condition and a drift is ascertained when it is severe enough or occurs for a period of time. A wide range of performance metrics can be used to assess the existence of drift in data streams. Referring to original work [27], two performance metrics, namely moving average and weighted moving average, are put forward. Since the moving average is more sensitive than the weighted version to concept change and thus being suitable in detecting abrupt drifts, it is used here and has the form $\widehat{X}_t = \sum_{t=1}^{\mathrm{P}} \Upsilon_t X_t, \Upsilon_t = 1/\mathrm{P}, \widehat{X} = \overline{X}$. Note that this can be calculated recursively with ease. The weighted version, conversely, is more reliable than unweighted version in detecting gradual drift. This approach is similar to the idea of statistical process control [53] except the basis of normality is relaxed here. Moreover, the use of the standard deviation $\sigma$ for the confidence interval is replaced by the significance level $\alpha$ which corresponds to the warning level ($\alpha_W$) and to the drift level ($\alpha_D$). The drift detection method is elaborated in Algorithm 2.

**Algorithm 2:** Drift Detection Method Based on the Hoeffding's inequality

*Input*

$\alpha_W \in (0,1], \alpha_D \in (0,1]$ are confidence for the warning level and the drift level

$\widehat{X}_{cut}$ is statistic computed from $x_1, x_2, \ldots, x_{cut}$

$\widehat{Y}_{t-cut}$ is statistic computed from $x_{cut+1}, \ldots, x_{\mathrm{P}}$, $\widehat{Z}_{\mathrm{P}}$ is statistic computed from $x_1, x_2, \ldots, x_{\mathrm{P}}$

$\varepsilon_{\widehat{X}_{cut}}, \varepsilon_{\widehat{Y}_{\mathrm{P}-cut}}, \varepsilon_{\widehat{Z}_{cut}}$ are respectively error bounds in accordance with statistics used

A data chuck $D = [x_1, \ldots, x_t, \ldots, x_{\mathrm{P}}] \in \Re^{\mathrm{P}\times n}$ containing $\mathrm{P}$ samples is received

*Output*

$State \in \{Stable, Warning, Drift\}$

Calculate the statistics $\widehat{Y}_{t-cut}, \widehat{Z}_{\mathrm{P}}$ and the error bounds $\varepsilon_{\widehat{Y}_{t-cut}}, \varepsilon_{\widehat{Z}_t}$ using the newest observation $x_t$ // calculate statistics of three data partitions and confidence intervals

**IF** $\widehat{Z}_t + \varepsilon_{\widehat{Z}_t} \le \widehat{X}_t + \varepsilon_{\widehat{X}_t}$

$\widehat{X}_{cut} = \widehat{Z}_t, \varepsilon_{\widehat{X}_{cut}} = \varepsilon_{\widehat{Z}_t}$, reset $\widehat{Y}_{t-cut}, \varepsilon_{\widehat{Y}_{t-cut}}$ // find the cut points

**End IF**

**IF** $H_0 : E[\widehat{X}_{cut}] \le E[\widehat{Y}_{t-cut}]$ is rejected with significance level $\alpha_D$ // determine the current state of data streams

$State \leftarrow Drift$, create a new classifier based on a current data chunk

**ElseIF** $H_0 : E[\widehat{X}_{cut}] \le E[\widehat{Y}_{t-cut}]$ is rejected with size $\alpha_W$

$State \leftarrow Warning$, do nothing but store current data chunk in the buffer prepare a new classifier if a drift is confirmed

**Else** $H_0 : E[\widehat{X}_{cut}] > E[\widehat{Y}_{t-cut}]$

$State \leftarrow Stable$, use data chunk to train a winning classifier and clear data samples in the buffer **End**

It is observed from Algorithm 2 that a new classifier is created when the drift state is signaled and is constructed using a current data chunk only. A transition period from warning to drift is required to bear out whether a change really occurs and is not caused by noise or outliers. No buffer is deployed to accumulate data in the transition period (warning to drift) to prevent a mixed-up concept of a new classifier. First, we start by finding a cut point in the current chunk which indicates a point where a population mean increases. The cut point is a switching point when $\widehat{Z}_t + \varepsilon_{\widehat{Z}_t} \le \widehat{X}_t + \varepsilon_{\widehat{X}_t}$ where $\widehat{X}_t, \widehat{Z}_t$ are statistics obtained from $x_1, x_2, ..., x_{cut}$ and $x_1, x_2, x_{cut}, .., x_{P=cut+m}$ respectively, while the error bounds $\varepsilon_{\widehat{X}_{cut}}, \varepsilon_{\widehat{Z}_{cut}}$ are calculated as follows:

$$\varepsilon_\alpha = (b-a)\sqrt{\frac{(m)}{2cut(m+cut)}\ln(\frac{1}{\alpha})} \quad (1)$$

where *a, b* are the minimum and maximum values of an input variable $[a,b]$. $\alpha$ is the significance level. After finding the cutting point, data points in the chunk are grouped in two groups $\widehat{X}_{cut} = [x_1, x_2, ..., x_{cut}], \widehat{Y}_{t-cut} = [x_{cut+1}, x_{cut+2}, ..., x_P]$. The two groups are used in the analysis of the null hypothesis to examine the current state of data streams. When a null hypothesis is valid, no change is detected in the current data stream. When the null hypothesis is rejected with the size $\alpha_W$, the warning status is reported but when it is rejected with the size $\alpha_D$, the drift status is returned. The null hypothesis is formulated as $H_0 : E(\widehat{X}_{cut}) \le E(\widehat{Y}_{t-cut})$ and its alternative is set as $H_1 : E(\widehat{X}_{cut}) > E(\widehat{Y}_{t-cut})$. The condition to reject the null hypothesis is set as $\widehat{X}_{cut} - \widehat{Y}_{t-cut} \ge \varepsilon_\alpha$ where $\varepsilon_\alpha$ is elicited using (1). We apply the same settings in [27] where $\alpha_W, \alpha_D$ are respectively fixed at 0.005 and 0.001. It is worth stressing that these two values has clear statistical interpretation because it represents the confidence level of the Hoeffding's bound in the level of $1-\alpha$.

It is observed in Algorithm 2 that no learning scenario is carried out at the warning stage. This mechanism is chosen since the warning phase constitutes a transition period where the presence of concept drift still calls for further investigation. During this phase, current data chunk is stored in the data buffer and is meant to construct a new classifier when a drift is signaled. This is, however, reset and all data are cleared from the buffer provided the next observation returns to the stable phase. The stable phase implies that the concept remains the same and does not induce an introduction of a new classifier. It, however, calls for the winning classifier to be updated using current data chunk to assure generalization's capability of the ensemble classifier because it reduces the risk of overfitting by feeding more observations to the base-classifiers. The winning classifier is selected by simply inspecting its predictive error - Mean Square Error is used in pENsemble.

*2) Ensemble Pruning Strategy Based on Local Generalization Error*: The success of ensemble classifier is highly determined by the generalization potential of base classifiers. Although it is well-known that a collection of weak classifiers often promotes better performance than that of strong classifiers, it is not the case in realm of data streams. The diversity comes at the cost of complexity and predictive performance because data stream is inherent with non-stationary contexts. A base classifier with low generalization potential is expected to play little during its lifespan or even to jeopardize final predictions because of their roles in the voting process and therefore pruning such base classifier reduces the ensemble complexity [22]. Our approach is inspired by the localized generalization error method which quantifies generalization capability of a classifier within a predefined Q local region [28]. This technique is meant to approximate the upper bound of mean square error (MSE) for unseen samples lying in the Q region. The use of a predetermined Q region is a plausible approach to study model's generalization since most training samples occupy a dense local region and are inter-related to each other because they are drawn from the same unknown distribution. Finding an upper bound of generalization error for hidden context in the entire input space is extremely difficult but we can safely ignore irrelevant concept sitting far away from training samples.

The Q neighbourhood is defined as that $S_Q(x_b) = \left\{ x \,\middle|\, x = x_b + \Delta x_i, 0 < |\Delta x_i| \le Q, i = 1,..,n \right\}$ where *n* is the number of input dimension and *Q* is a given real value [28]. All samples in $S_Q(x_b)$ except $x_b$ are regarded as unseen samples and the generalization capability of a model must be delved from its generalization capability in a union of $S_Q(x_b)$. Since a complete picture of data distribution is unknown before the process runs, it is assumed that unseen samples have same chance to appear. $\Delta x_i$ is treated as a random variable following the uniform distribution with zero mean and variance $\sigma^2_{\Delta x_j}$. The localized generalization error is defined as follows:

$$R_{SM}(Q) = \int_{S_Q} (f_i(x) - F(x))^2 p(x) dx \quad (1)$$

where $f_i(x)$, $F(x)$, $p(x)$ are the *i-th* local expert, the target function and the unknown probability density function of the input $x$ respectively. In practice, unseen samples will lead to a higher error than those of training samples. Through the Hoeffding's inequality with a probability of $(1-\eta)$, the average of the square error converges to the true mean:

$$R_{SM}(Q) \le (\sqrt{R_{emp}} + \sqrt{E_{SQ}((\Delta y)^2)} + A)^2 + \varepsilon \quad (2)$$

$$\Delta y = f(x) - f(x_b),\ \varepsilon = B\sqrt{\ln \eta / (-2P)},\ R_{emp} = \frac{\sum_{t=1}^{P} (f_i(x_b) - F(x_b))^2}{P}$$

where $A, B, P, \eta$ stand for the difference between the maximum and minimum values of the desired outputs, the maximum possible value of the MSE, the data window size, and the confidence interval. The range of desired output, $A$, and the maximum MSE, $B$, are known during the training process and are updated regularly as new training samples are observed . $R_{emp}$ denotes the training error which indicates the bias of a model. $E_{SQ}((\Delta y)^2)$ stands for the stochastic sensitivity measure which illustrates the sensitivity of network output against the variation of network input.

The difference between the training sample and the unseen sample within the Q neighbourhood is portrayed by the output perturbation $\Delta y$ and $E_{SQ}((\Delta y)^2)$ indicates the expectation of the squared output perturbations between already seen samples and unknown samples in the Q local region. It analyses how sensitive a classifier's output is to the variation of input data. The expression of the stochastic sensitivity measure for a Gaussian basis function with a center $u_j$ and a width $v_j$ of *j-th* input coordinate has been defined in [28] and is formulated by assuming independent input perturbations without the weight perturbations. The input perturbation follows the uniform distribution with zero mean and a variance $\sigma^2_{\Delta x_j}$ but the input feature is not identically distributed and has its own expectation $\mu_{x_j}$ and variance $\sigma^2_{x_j}$. The definition of the stochastic sensitivity measure is applicable [29] directly here when pClass is implemented with a classical axis-parallel rule. This strategy can be still used for the generalized Gaussian rule if a transformation strategy is carried out to find low dimensional representation of multivariate Gaussian function [47]. By the central limit theory, if the number of input features is not too low, the Gaussian basis function would have a log-normal distribution. It is written as follows:

$$E_{SQ}((\Delta y)^2) = \frac{1}{\mathrm{P}}\sum_{t=1}^{\mathrm{P}} \int_{S_Q(x_b)} (f(x_b+\Delta x) - f(x_b))^2 p(\Delta x) d\Delta x \quad (3)$$

where $p(\Delta x)$ stands for the probability density function of the input perturbation. Since the input perturbation is uniformly distributed in the Q region, the probability density function is formed as $1/(2Q)^n$ and the variance is expressed as $\sigma^2_{\Delta x_j} = Q^2/3$. The assumption of uniformly distributed input perturbations is plausible considering the strictly single-pass working principle of pENsemble without any prior knowledge. Albeit this assumption, the distribution of the input perturbations can be relaxed provided that the variance of the input perturbation is finite. Let $\varphi_i = (x_e^T W_i)\exp((Var(s_i)/2v_i^4) - (E(s_i)/v_i^2)), s_i = \left\| x - u_j \right\|^2$,

$$E(s_j) = \sum_{j=1}^{n} (\sigma^2_{x_j} + (\mu_{x_j} - u_{ij})^2), \varsigma_i = \varphi_i / v_i^4$$

$$Var(s_i) = \sum_{j=1}^{n} \begin{matrix} (E_D[(x_j - \mu_{x_j})^4] - (\sigma^2_{x_j})^2 + 4\sigma^2_{x_j}(\mu_{x_j} - u_{i,j})^2 \\ +4E_D[(x_j - \mu_{x_j})^3](\mu_{x_j} - u_{i,j})) \end{matrix}$$

$$v_i = \varphi_i (\sum_{j=1}^{n} (\sigma^2_{x_j} + (\mu_{x_j} - u_{i,j})^2 / v_i^4))$$

The final expression of the stochastic sensitivity measure $E_{SQ}((\Delta y)^2)$ [28]-[31] is formulated in the form:

$$E_{SQ}((\Delta y)^2) \approx \sum_{i=1}^{R} \varphi_i ((\sum_{j=1}^{n} \sigma^2_{\Delta x_j} (\sigma^2_{x_j} + (\mu_{x_j} - u_{i,j})^2 + 0.2\sigma^2_{\Delta x_j}) / v_i^4) = \frac{1}{3} Q^2 \sum_{i=1}^{R} v_i + \frac{0.2 Q^4 n}{9} \sum_{i=1}^{R} \varsigma_i \quad (4)$$

Because $Q$ is constant for all base-classifiers, it can be dropped from (6). It is observed from the localized generalization error formula (2) that there exist three components: the training error, the stochastic sensitivity measure and some constants. High training error pinpoints the under-training case which results in poor generalization of unseen samples. The stochastic sensitivity measure illustrates the sensitivity of a classifiers against output's change and that having its outputs varying dramatically against input variation should characterize high stochastic sensitivity. A good generalization is attained by minimizing both terms or forming a sound tradeoff between the two. In other words, the ensemble pruning scenario aims to discover those classifiers with large $R_{SM}(Q)$ because the smaller it is, the better the model's generalization is. Although this formula aims to analyse the upper bound of MSE which targets regression cases and direct regression to class indices in most cases results in poor performance, this strategy is still applicable to classification problems. The relationship between the localized generalization error and misclassification rate has been studied in [31] where if the error distribution is known, the percentage of unseen samples being correctly classified is given by $0 \le |E(err)| \pm \gamma\sqrt{Var(err)} \le 0.5$ where $\gamma$ is the confidence parameter. Suppose that we compare two classifiers $f_1, f_2$, it is understood from the localized generalization error theory that $f_1$ is said to have better generalization error when its $R_{SM}(Q)$ is lower than that of $f_2$ with the same $Q$. It is shown in [31] that the generalization performance of $f_1$ in terms of misclassification rate is better than $f_2$ with the minimum probability $(1 - \frac{1}{6}\sqrt{R_{SM}(Q)_1 / R_{SM}(Q)_2})(1-\eta)^3$, where $(1-\eta)$ is the confidence level.

The ensemble pruning condition is set as follows:

$$R_{SM}(Q)_i < mean(R_{SM}(Q)_i) - 3std(R_{SM}(Q)_i) \quad (5)$$

where this expression adopts the 3-sigma rule principle and aims to track downtrend of the model's generalization. Assuming that the localized generalization error follows the Gaussian distribution, 99.7% of its values occupy the three sigma range or it incurs 99.7% confidence level. That is, any case beyond the range of three sigma is said to be anomalies. Although the concept of localized generalization error has been exploited in various problems [28]-[31], its efficacy for data stream analytics is to the best of our knowledge unexplored.

The ensemble pruning strategy has drawn significant research attention to enhance scalability of ensemble learner in the large-scale applications since the increase in complexity of the ensemble classifier is not obvious in small-scale applications. In [73], the notion of margin distribution with $L_1$ norm regularization strategy is implemented to produce sparse weight for ensemble pruning purpose. Similar idea is adopted in [74] but in addition to ensemble margin the confidence concept is incorporated. Another prominent work for ensemble pruning is shown in [75]. It relies on a combination of several diversity measures and the use of Genetic Algorithm (GA) to weight them. The individual contribution ordering based method for ensemble pruning scenario is proposed in [76] and aims to find diversity/accuracy tradeoff. The expectation propagation is used to derive sparse weights for ensemble pruning method in [77]. While the ensemble pruning scenario is a mature research area, vast majority of them are computationally prohibitive for data streams applications. Moreover, few works have considered generalization potential for the underlying indicator of ensemble pruning as exemplified by pENsemble.

*3) Online Feature Selection Strategy*: A high input dimension is commonly found in various real-world data stream cases and undermines the learning capability in the online real-time scenario because it imposes considerable complexity [33]. The transparency of a fuzzy rule is also affected because a rule consists of too many atomic clauses. Notwithstanding that the online feature selection strategy has drawn considerable research interest, they to date focus on a single classifier only. An online feature selection technique for the ensemble learner is proposed in this paper and is constructed under the framework of the GOFS method [32]. As the OFS [33], our feature selection approach is extensible to the partial input information condition where only a subset of input attributes can be obtained for the training process. The GOFS performs a crisp feature selection where input features are assigned crisp weights (0 or 1) which allows dynamic activation and deactivation of input attributes during the training process.

The contribution of *j-th* input features can be measured from an accumulated output weight across all fuzzy rules $\sum_{i=1}^{R} W_{j,i}$ because it indicates how much output change is imposed by a variation of input attributes [2]. Since pENsemble is developed from a collection of first order TSK fuzzy systems $W_i \in \Re^{(n+1)\times 1}$ , the 0-th term of the first order TSK fuzzy system, which corresponds to the intercept of a linear function, is excluded from the summation of output weights. In realm of the TSK fuzzy system, the rule consequent depicts the local tendency of a rule and may substitute the gradient information in the sensitivity analysis of input variables since the gradient information changes in each point in the case of nonlinear function. This concept is confirmed by the fact that each base classifier in the pENsemble employs a local learning scheme in which each rule consequent represents a specific region of the approximation curve. Data standardization is required here because different input ranges may mislead the contribution of an input feature. To guarantee transparency of feature contribution, normalization is done:

$$\kappa_j = \frac{\sum_{i=1}^{R} W_{j,i}}{\sum_{j=1}^{n}\sum_{i=1}^{R} W_{j,i}} \tag{8}$$

where $\kappa_j$ is the contribution of *j-th* input attribute. Since pENsemble consists of a set of evolving classifiers, fuzzy rules of all local experts are extracted and subject to (8) where $R = R_1 + R_2 + ... + R_M$ is a total number of fuzzy rules of all base classifiers while $M$ is the number of base classifiers in the ensemble. In addition, a sparsity property of L1 norm is examined to understand whether the value of *n* input features is accumulated in the L1 ball. Referring to the OLS theory, the input pruning process takes place given that misclassification occurs. The input pruning scenario is executed here when the global prediction of ensemble network does not match the true class label $C \neq \hat{C}$ where $C$ is the true class label and $\hat{C}$ is the predicted class label. This approach is plausible because the feature selection scenario aims to take the corrective actions by getting rid of the influence of poor features. No feature selection is necessary when correct prediction is returned to save computational cost. The rule consequent is first adjusted using the gradient descent approach and projected to the L2 ball to assure a bounded norm. Detailed procedure of the GOFS method is shown in Algorithm 3.

| **Algorithm 3:** GOFS procedure for full input attributes |
|---|
| *Input*: $\alpha, \chi, B$ are the learning rate, the regularization factor and the desired number of input dimension.<br>*Output*: $X_{selected} \in \Re^{1\times B}$ is a selected input vector |
| Obtain the global prediction of the ensemble network $\hat{C}$<br>**IF** $C \neq \hat{C}$<br>// update the rule consequent of all base classifiers $W_i = W_i - \alpha\chi W_i - \alpha\chi \frac{\partial E}{\partial W_i}$ .<br>//Project the weight vector into the L2 ball $W_i = \min(1, \frac{1/\sqrt{\chi}}{\lVert W_i \rVert_2}) W_i$<br>// Compute the contribution of input attributes as per (1)<br>// Extract $X_{selected}$ from the highest $B$ elements of (1)<br>**Else**<br>$W_i = W_i - \alpha\chi W_i$<br>**End IF** |

We fix $\alpha = 0.2, \chi = 0.01$ following the same setting as [32]. The standard mean square error (MSE) is applied as the cost function, the first order derivative $\frac{\partial E}{\partial W_i}$ is derived as follows:

$$\frac{\partial E}{\partial W_i} = -\frac{\sum_{i=1}^{R} x_e \varphi_i}{\sum_{i=1}^{R} \varphi_i} \tag{2}$$

where $\varphi_i$ is the spatial firing strength. It is worth noting that (2) is elicited under assumption that all fuzzy rules are structured under the first order TSK fuzzy neural network under pClass framework. In other words, fuzzy rules of all base classifiers are combined and treated as a unified local expert. This scenario is made possible by the local property of the pENsemble where each fuzzy rule functions as a loosely coupled sub-model. The stochastic gradient descent approach is applied in Algorithm 2 rather than the FWGRLS method because no covariance matrix has to be allocated and assigned for each local model thereby greatly simplifying the overall optimization process. It is worth noting the feature selection process is done in a centralistic manner where all fuzzy rules of each base classifiers are put together. Hence, the output covariance matrix in the local level cannot be used as it represents different optimization objectives. The convergence of the GOFS method has been proven [32] and its upper bound has been obtained. The GOFS method allows different subsets of input variables to be selected in every training observation. Since the partial input information situation only entails minor variation of its full counterpart [32], it is not explained here.

*4) Complexity Analysis*: This section aims to analyze the computational burden of pENsemble which presents a generalized version of DWM. The pENsemble utilizes the drift detection method which imposes the computational complexity $O(n)$ because it only relies on the mean of data samples which can be computed with ease recursively. The computational complexity of pENsemble is also affected by the rule pruning scenario governed by the localized generalization error. This learning module incurs the computational burden $O(nRO)$ for one classifier. Suppose that there exist $M$ classifiers in the classifier, this figure increases to $O(nROM)$. The resultant computational complexity of pENsemble is $O(M + P(DDM + EP) + IP)$ where $DDM$ stands for the drift detection method, $EP$ denotes the ensemble pruning and $IP$ is a short of the input pruning. $P$ is the data chunk size and data samples in the chunk are learned in a single scan and are not revisited again. Note that the term $M$ in the aforementioned big $O$ notation is influenced by the computational complexity of pClass as the base classifier. pClass is a fully evolving algorithm working in the single-pass learning mode. The computational complexity of pClass has been derived in [24].

## IV. Parsimonious Classifier

This section briefly outlines algorithmic procedure of pClass which serves as the local expert of pENsemble. It includes network structure of pClass, rule growing strategy, rule pruning and recall strategy, and parameter learning strategy. Since pENsemble deploys the online feature selection scenario in the top level, the input weighting mechanism of pClass is switched to the sleep mode.

- *Network Structure of pClass*: pClass is a class of neural-fuzzy classifiers generating a generalized first-order TSK fuzzy rule. It utilises a multivariate Gaussian function evolving a non-axis-parallel ellipsoidal cluster as the rule premise, while exploiting the first order polynomial as the rule consequent. The multivariate Gaussian function offers an appealing input space partition notably when data are not distributed in the underlying axes because the ellipsoidal cluster rotates to any direction [24]. Such trait is capable of lowering the fuzzy rule demand and retains inter-relations among input variables [11]. Although such rule premise is less transparent than the conventional fuzzy rule, pClass is fitted with a transformation strategy which allows the extraction of classical rule. Nonetheless, the multivariate Gaussian function incurs prohibitive computational cost and memory requirement due to its non-diagonal covariance matrix. pClass is also implemented here using the conventional Gaussian rule triggering axis-parallel cluster. This also aims to demonstrate the performance of pENsemble under two different base classifiers.
- *Rule Growing Strategy*: the rule growing process of pClass is orchestrated by three rule growing modules which determines the novelty of a data point whether it deserves to be a prototype of a new rule. The first rule growing strategy, namely the Datum Significance (DS) method, estimates the statistical contribution of a data sample which indicates its possible contribution in the whole course of training process. It is derived from assumption of the uniform distribution and the statistical contribution is expressed as the zone of influence of an ellipsoidal cluster.

The statistical contribution, however, ignores summarization power of a rule because it does not consider how strategic a current position of rule in the feature space is [24], [38]. This hinders its capability to capture concept drift because no distance information is provided in enumerating the importance of fuzzy rules. The second rule growing strategy, namely the Data Quality (DQ) method, is put forward. This concept follows the concept of recursive density estimation (RDE) [2], [7] where a density of a local region is computed recursively. This concept concludes that a rule addition is necessary either when a data point represents the most relevant concept having the highest density or when a data point is beyond the coverage of existing rules [24]. The DQ method differs from the RDE method [7] in two facets: 1) it involves a weighting strategy reducing the influence of outliers which causes a drop of density for next samples; 2) it uses the inverse multi-quadratic function in lieu of the Cauchy function; 3) it is tailored for the multivariate Gaussian function.

An oversized rule is prone to the cluster delamination problem which pinpoints a situation where two or more distinct data clouds [2] are contained by only a cluster. This situation undermines the generalization because the specificity of a cluster decreases significantly. The third rule growing strategy aims to overcome this issue borrowing the concept of GART+ [54]. It monitors the coverage span of the winning rule obtained from the Bayesian concept – a rule with the maximum posterior probability. It limits the growth of the winning rule where a new rule is introduced when the size of winning rule exceeds a pre-specified level [55].

- *Rule Pruning and Recall Strategy*: pClass is equipped by two rule pruning strategies, namely extended rule significance (ERS) method, and potential+ (P+) method. The ERS method shares the same principle of the DS method which estimates the statistical contribution of fuzzy rules to discover inconsequential rules which play little role to the final output during their lifespan. It combines significance of both rule premise and rule consequence to quantify the rule contribution. The significance of rule premise is derived from the approximation of accumulated contribution of the multivariate Gaussian function during its lifespan without revisiting preceding samples. It is obtained under a uniform distribution assumption and this assumption results in a zone of influence of fuzzy rules as an indicator of rule premise significance. The contribution of rule consequent is measured from a weighted sum of an output weight vector since a small rule weight normally generate negligible outputs.

The P+ method monitors the evolution of a rule in respect to current data trend and is vital in non-stationary environments. It aims to find obsolete rules which are no longer relevant to delineate recent concept due to drift. This scenario is realised by extending the concept of data potential [7], [56] for the rule pruning scenario. The concept of data potential performs recursive density estimation of fuzzy regions which pinpoints relevance of fuzzy rules since fuzzy rules which are not supported by current data distribution is expected to return low density. The P+ method, however, differs from the data potential concept in its kernel function using the inverse multi-quadratic function instead of the Cauchy function. The P+ method also functions as the rule recall scenario which is capable of handling the recurring drift. That is, the recurring drift refers to a situation where previous data distribution reappears again in the future. This may trigger previously pruned rules portraying old concept to be valid again. Adding a completely new rule to address the cyclic drift does not coincide with the flexible nature of human being which can recall previous knowledge with ease. Furthermore, adding a new rule risks on catastrophic forgetting of previously valid knowledge because it ignores learning history. Previously pruned rules can be reactivated in the future provided that its relevance indicated by the P+ method beats existing rules and newly observed data point. It is worth noting that previously pruned rules are discounted from any training scenarios except the update of their densities. This paradigm ensures that the rule pruning scenario still relieves the computational burden.

- *Parameter Learning Strategy*: Data streams may not incur sufficient novelty to be a prototype of a new rule but such data streams are useful to refine the influence zone of existing rule base [24]. This situation is addressed by fine-tuning the rule premise of the winning rule. The adaptation scenario is derived from the sequential version of maximum likelihood and is adapted to the multivariate Gaussian function. Furthermore, pClass utilises a direct update scheme of the inverse covariance matrix according to the formulas derived in [39] which shelves the reinversion of the covariance matrix. The winning rule is determined using the Bayesian concept where a rule with the maximum posterior probability is selected as the winning rule. This winning rule selection is preferred over the compatibility measure [55] since it takes into account the rule's population.

The rule consequent is adjusted using the fuzzily weighted generalized recursive least square (FWGRLS) method. The FWGRLS is a derivation of the FWRLS method [7] borrowing the concept of weight decay function of the GRLS method in [57]. The FWGRLS method works in the local learning scenario well-suited to the EIS since it offers a decoupled adaptation scheme where adaptation of each local region incurs no cross correlation to each other since each local sub-model features a unique output covariance matrix [24]. Learning in a particular sub-model has no effect to the stability and convergence of other rules. The salient feature of the FWGRLS method compared to the FWRLS method lies in the generalized weight decay term in the cost function which aims to alleviate the overfitting situation. The weight decay term also supports compactness and parsimony of the rule base because it forces the rule consequent of an inconsequential rule to a small range. Therefore, inconsequential rules can be located by the ERS method easily. The quadratic weight decay term is incorporated since it is capable of reducing the weight vector proportionally to its current values [47].

## V. EXPERIMENTAL STUDIES

We elaborate on numerical validations of pENsemble by using 15 real-world data streams and comparisons with prominent classifiers. The simulations were undertaken with an Intel (R) Core i5-6600 CPU @ 3.3 GHZ with 8 GB of RAM. pENsemble is implemented under MATLAB environment where the MATLAB code is made publicly available in[1)].

### A. Comparisons with State-of-The Art Algorithms

pENsemble is benchmarked against four prominent classifiers for data streams. pENsemble is realized under two versions of pClass: axis-parallel, multivariate. The underlying feature of consolidated algorithms are elaborated as follows:

- *Learn++NSE* is seen as one of pioneer works in dynamic ensemble classifier for non-stationary environments [18]. It presents an extension of Learn++ [58] to tackle concept drifts in data streams. It is an Adaboost-like algorithm which consists of a set of weak learners and adopts the concept of sample weighting. The underlying contribution is observed in the dynamically weighted majority voting which reflects dynamic contexts of data streams.
- *Learn++CDE* is a generalized version of Learn++NSE integrating a specific mechanism to handle the class imbalanced problem in data streams [52]. It combines the Learn++NSE with the well-established SMOOTE using the concept of undersampling and oversampling approaches for imbalanced data streams. It also proposes concepts of subensemble and class independent error weighting with a penalty constraint. Both Learn++NSE and Learn++CDE make use of CART as the base classifier.
- *pClass* is a class of evolving classifier putting forward the open paradigm and the online learning capability [24]. pClass is structured as a five-layered neural network working in tandem and actualising a generalized TSK fuzzy inference system. In addition to its flexible network structure, pClass is equipped by an online feature weighting strategy. All of which are summed up in Section 4 of this paper. This comparison is necessary to illustrate how the proposed ensemble learning scheme is better than its single classifier version.
- *eT2Class* is another case of evolving classifiers unifying the dynamic network structure and the online learning capability [10]. It differs from pClass since it incorporates the interval type-2 fuzzy working principle. It features a fast type-reduction method which is scalable for the online data stream processing.

Table 1. Characteristic of data streams

| *Data stream* | *IA* | *C* | *DP* | *TS* | *TRS* | *TES* |
|---|---|---|---|---|---|---|
| *SEA* | *4* | *2* | *100000* | *200* | *250* | *250* |
| *Iris+* | *4* | *4* | *450* | *10* | *34* | *11* |
| *Car+* | *6* | *2* | *1728* | *10* | *130* | *42* |
| *Electricity pricing* | *8* | *2* | *45312* | *200* | *150* | *77* |
| *Weather* | *4* | *2* | *60000* | *10* | *1000* | *5000* |
| *Line* | *2* | *2* | *2500* | *10* | *200* | *50* |
| *Sin* | *2* | *2* | *2500* | *10* | *200* | *50* |
| *Sinh* | *2* | *2* | *2500* | *10* | *200* | *50* |
| *10dplane* | *10* | *2* | *1200* | *10* | *100* | *20* |
| *Gaussian* | *4* | *2* | *800K* | *100* | *400* | *7200* |
| *Hyperplane* | *4* | *2* | *120 K* | *100* | *1000* | *250* |
| *Susy* | *18* | *2* | *5 M* | *10000* | *400* | *100* |
| $CHD_1$ | *10* | *2* | *965* | *5* | *100* | *93* |
| $CHD_2$ | *10* | *2* | *528* | *5* | *100* | *28* |

IA: Input Attributes, C: Classes, DP: Data Points, TS: Time Stamps, TRS: Training Samples, TES: Testing Samples

Consolidated algorithms were numerically validated using 15 real-world and synthetic data streams featuring highly dynamic characteristics. Popular artificial and semi-artificial Diversity for Dealing with Drifts (DDD) problems characterizing the abrupt and gradual drifts, namely sin, sinh, line, 10dplane, car and iris+, were explored to investigate the

1) http://www.ntu.edu.sg/home/mpratama/Publication.html

performance of consolidated algorithms [59], [60]. The SEA problem introduced in [61] was used to bear out the efficacy of benchmarked algorithms. Moreover, an extension of the SEA problem contributed by Ditzler and Polikar [52] was put forward instead of its original version since it offers the class imbalance property and the cyclical drift which often occurs in the real-world data streams. Another popular problem in the data stream mining area, namely the Gaussian problem, was exploited [18] where each class contains gradual and independent drift which can be controlled from the mean and variance of the parametric equations. The hyperplane problem was exploited to inspect the learning performance of consolidated algorithms - a benchmark problem in the massive online analysis (MOA). To assess scalability of consolidated algorithms, a big dataset with 5 million records - SUSY- was utilised. On top of those artificial and semi-artificial data streams, six real world problems, namely electricity pricing, weather, coronary heart disease for male and female patients (Courtesy of Dr. Agus Salim, Latrobe) [63] were included in our experiments. The characteristics of these data streams along with experimental procedures are encapsulated in Table 1 and detailed descriptions of each problem is detailed in the supplemental document.

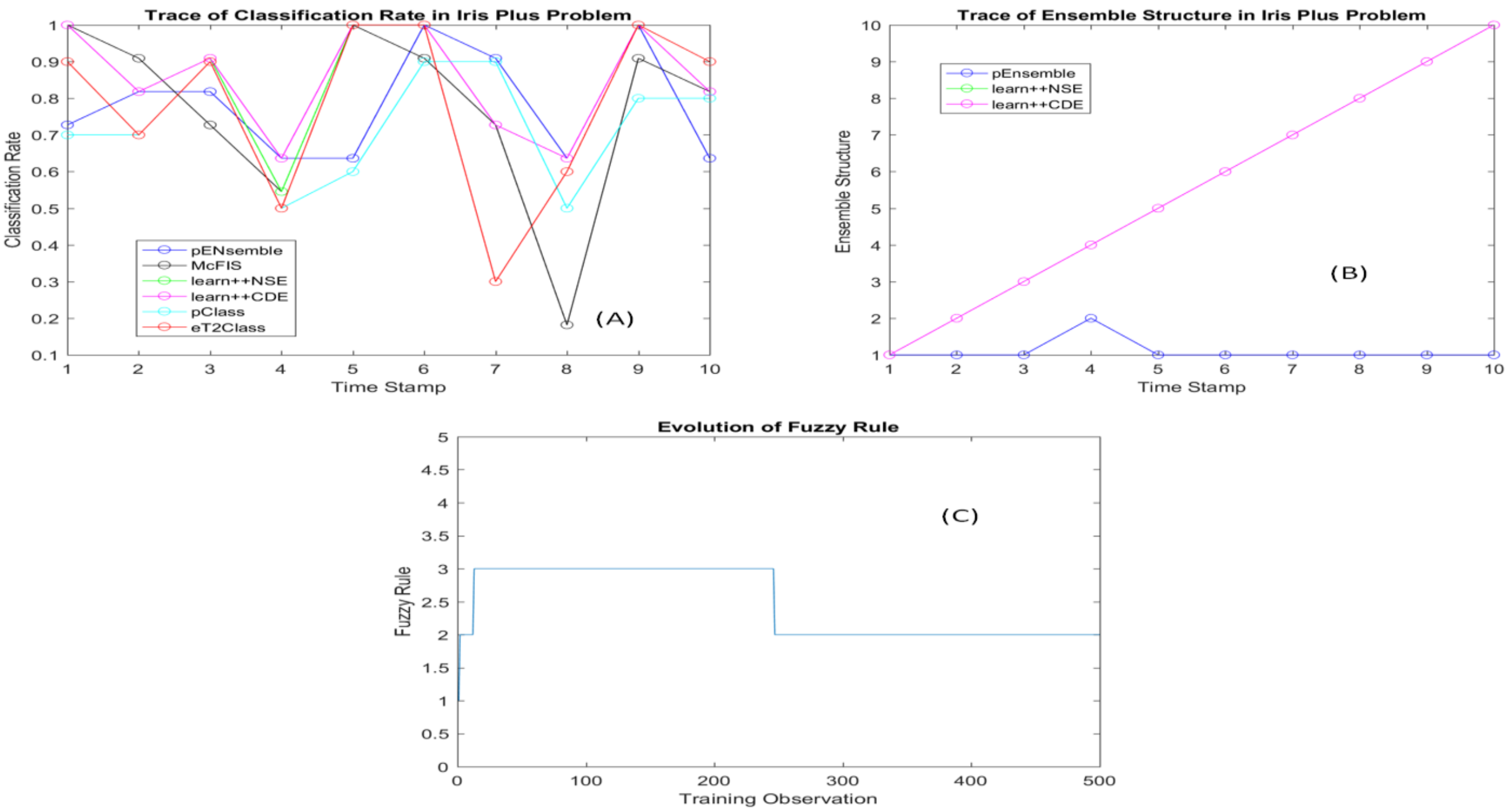

Fig. 2(a) trace of classification rate in iris plus problem, (b) trace of ensemble structure in iris plus problem, (c) fuzzy rule evolution of base classifier

Table 2. Numerical results of benchmarked algorithms

| Numerical Example | Evaluation Criteria | pENsemble (axis-parallel) | Learn++.NSE | Learn++.cde | pClass | eT2Class | pENsemble (multivariate) |
|---|---|---|---|---|---|---|---|
| SEA | Classification Rate | **0.97±0.02** | 0.93±0.02 | 0.93±0.02 | 0.89±0.1 | 0.88±0.23 | **0.97±0.02** |
| | Fuzzy Rule | 4.1±1.8 | N/A | N/A | 6.6±4.2 | **1.5±0.5** | 4.06±1.8 |
| | Input Attribute | **2** | 3 | 3 | 3 | 3 | **2** |
| | Network Parameters | **40.6±18.3** | N/A | N/A | 157.3±101.9 | 61.3±21 | 48.7±21.8 |
| | Execution Time | 0.78±0.2 | 1804.2 | 2261.1 | 0.42±0.3 | **0.34±0.11** | 0.89±0.12 |
| | Ensemble Size | **2.03±0.9** | 200 | 200 | N/A | N/A | **2.03±0.3** |
| Line | Classification Rate | 0.9±0.07 | 0.88±0.13 | 0.89±0.14 | 0.91±0.07 | **0.94±0.1** | 0.9±0.07 |
| | Fuzzy Rule | 2 | N/A | N/A | 1.5±0.7 | **1.1±0.3** | 2 |
| | Input Attribute | **1** | 2 | 2 | 2 | 2 | **1** |
| | Network Parameters | **12** | N/A | N/A | 30 | 22 | **12** |
| | Execution Time | 0.36±0.14 | 1.24 | 1.53 | 0.25±0.009 | 0.13±0.04 | **0.06±0.06** |
| | Ensemble size | **1** | 10 | 10 | N/A | N/A | **1** |
| sin | Classification Rate | 0.78±0.26 | 0.8±0.15 | **0.8±0.13** | 0.72±0.2 | 0.71±0.3 | 0.78±0.3 |
| | Fuzzy Rule | 2.9±1.6 | N/A | N/A | 3.3±1.2 | **1.9±1.1** | **2.8±1.4** |
| | Input Attribute | **1** | 2 | 2 | 2 | 2 | 1 |
| | Network Parameters | 17.4±9.9 | N/A | N/A | 39.6 | **38±11.3** | **16.8±8.4** |
| | Execution Time | 0.21±0.05 | 0.8 | 1.9 | **0.17±0.04** | 0.24±0.03 | 0.19±0.05 |
| | Ensemble Size | 1.6±0.5 | 10 | 10 | N/A | N/A | **1.4±0.7** |
| sinH | Classification Rate | 0.71±0.06 | 0.73±0.22 | **0.75±0.5** | 0.71±0.09 | 0.69±0.06 | 0.71±0.06 |
| | Fuzzy Rule | 2 | N/A | N/A | 2±0.9 | **1.2±0.7** | 2 |
| | Input Attribute | **1** | 2 | 2 | 2 | 2 | 1 |
| | Network Parameters | 44 | N/A | N/A | 43.2 | **24±8.4** | 44 |
| | Execution Time | 0.18±0.09 | 0.69 | 1.89 | 0.12±0.05 | 0.18±0.05 | **0.19±0.04** |
| | Ensemble Size | **1** | 10 | 10 | N/A | N/A | 1 |
| | Classification Rate | 0.78±0.15 | 0.84±0.17 | **0.85±0.14** | 0.73±0.18 | 0.78±0.2 | 0.75±0.17 |
| | Fuzzy Rule | **2.5±1.1** | N/A | N/A | 4.6±1.9 | 3.3±0.5 | 2.2±0.6 |

| Iris+ | Input Attribute | **1** | 4 | 4 | 4 | 4 | 4 |
|---|---|---|---|---|---|---|---|
| | Network Parameters | 25±10.8 | N/A | N/A | 147±22.1 | 237.6±34.8 | 22±6.3 |
| | Execution Time | 0.32±0.2 | 0.26 | 0.44 | 0.27±0.01 | 0.42±0.4 | **0.06±0.03** |
| | Ensemble Size | **1.1±0.3** | 10 | 10 | N/A | N/A | 1.1±0.3 |
| Car | Classification Rate | **0.79±0.1** | 0.67±0.3 | 0.68±0.3 | 0.77±0.1 | 0.77±0.14 | **0.79±0.1** |
| | Fuzzy Rule | 3 | N/A | N/A | 2.5±0.8 | 1.5±0.5 | 1 |
| | Input Attribute | **1** | 6 | 6 | 6 | 6 | 1 |
| | Network Parameters | 18 | N/A | N/A | 140±47.6 | 156±54.6 | **6** |
| | Execution Time | 0.29±0.21 | 1.36 | 1.34 | 0.09±0.07 | 0.16±0.05 | **0.19±0.05** |
| | Ensemble Size | **1** | 10 | 10 | N/A | N/A | 1 |
| 10dplane | Classification Rate | 0.78±0.2 | 0.72±0.14 | 0.71±0.13 | 0.63±0.26 | 0.56±0.38 | **0.8±0.2** |
| | Fuzzy Rule | 6.4±2.1 | N/A | N/A | **3.1±0.87** | 3.2±1.4 | **2.8±0.6** |
| | Input Attribute | **5** | 11 | 11 | 11 | 11 | 5 |
| | Network Parameters | **18** | N/A | N/A | 37.2±10.5 | 956.8±418.1 | 117.6±46.7 |
| | Execution Time | 0.22±0.08 | 0.79 | 1.37 | 0.39±0.42 | 1.2±0.6 | **0.25±0.08** |
| | Ensemble Size | 1.6±0.52 | 10 | 10 | N/A | N/A | **1.5±0.5** |
| Weather | Classification Rate | **0.8±0.02** | 0.75±0.03 | 0.73±0.02 | 0.8±0.04 | 0.8±0.03 | 0.78±0.02 |
| | Fuzzy Rule | 4.6±2.9 | N/A | N/A | 2.3±0.5 | **2.3±0.3** | 3±1.05 |
| | Input Attribute | **2** | 8 | 8 | 8 | 8 | 1 |
| | Network Parameters | **60±28.2** | N/A | N/A | 226.8±95.6 | 391±81 | 18±6.3 |
| | Execution Time | 2.0±0.7 | 184.8 | 9.98 | 1.8±0.22 | 1.8±0.1 | **1.4±0.06** |
| | Ensemble Size | **4.6±2.9** | 10 | 10 | N/A | N/A | 1.5±0.5 |
| Gaussian | Classification Rate | 0.75±0.0 | **0.95±0.03** | **0.95±0.03** | 0.74±0.2 | 0.72±0.13 | 0.75±0 |
| | Fuzzy Rule | 49.78±28.89 | N/A | N/A | 2.1±0.3 | **1.4±0.5** | 99.56±57.3 |
| | Input Attribute | **1** | 2 | 2 | 2 | 2 | **1** |
| | Network Parameters | **298.7±171.9** | N/A | N/A | 50.2±6.9 | 35.5±12.4 | 597.36±343.9 |
| | Execution Time | 27.7±13.3 | 21020 | 79998 | 0.74±0.05 | **1.6±0.3** | 19.6±9.3 |
| | Ensemble Size | 49.8±28.7 | 100 | 100 | N/A | N/A | **49.7±28.7** |
| Hyperplane | Classification Rate | **0.92±0.02** | 0.91±0.02 | 0.9 | 0.91±0.02 | 0.89±0.1 | **0.92±0.02** |
| | Fuzzy Rule | 3.74±0.7 | N/A | N/A | 3.8±1.7 | **2.04±0.2** | 2.7±0.7 |
| | Input Attribute | **2** | 4 | 4 | 4 | 4 | **2** |
| | Network Parameters | 37.4±6.7 | N/A | N/A | 114.9±52.6 | 110.6±10.6 | **32.8±8.1** |
| | Execution Time | 0.9±0.2 | 926.04 | 2125.5 | 2.7±1.4 | 2.5±1.5 | **0.7±0.23** |
| | Ensemble Size | **1.87±0.34** | 100 | 100 | N/A | N/A | **1.87±0.34** |
| Electricity pricing | Classification Rate | **0.75±0.16** | 0.69±0.08 | 0.69±0.08 | 0.68±0.1 | 0.72±0.17 | **0.75±0.16** |
| | Fuzzy Rule | 3.02±1.1 | N/A | N/A | 3.5±2.4 | 4.6±1.3 | **1.01±0.07** |
| | Input Attribute | 2 | 8 | 8 | 8 | 8 | **1** |
| | Network Parameters | 36.8±13.04 | N/A | N/A | 226.8±95.6 | 778.6±20.8 | **6.03±0.4** |
| | Execution Time | 0.28±0.08 | 211.2 | 211.2 | 7.1±4.4 | 0.3±0.08 | **0.27±0.02** |
| | Ensemble Size | **1.3±0.6** | 119 | 119 | N/A | N/A | **1.01±0.07** |
| Coronary Heart Disease$_1$ | Classification Rate | 0.94±0.02 | 0.83±0.4 | 0.84±0.35 | 0.86±0.16 | 0.94±0.02 | **0.95±0.01** |
| | Fuzzy Rule | 2 | N/A | N/A | 6.4±5.8 | 3.8±0.4 | **1** |
| | Input Attribute | 2 | 10 | 10 | 10 | 10 | **1** |
| | Network Parameters | 24 | N/A | N/A | 844.8±773.1 | 957.8±112.8 | **6** |
| | Execution Time | **0.26±0.06** | 1.87 | 1.82 | 0.77±0.28 | 1.92±0.8 | 0.28±0.08 |
| | Ensemble Size | **1** | 5 | 5 | N/A | N/A | **1** |
| Coronary Hearth Disease$_2$ | Classification Rate | 0.93±0.09 | 0.81±0.4 | 0.81±0.43 | 0.92±0.4 | 0.92±0.01 | **0.97±0.07** |
| | Fuzzy Rule | 2 | N/A | N/A | **1** | 3.2±1.1 | **1** |
| | Input Attribute | 2 | 10 | 10 | 10 | 10 | **1** |
| | Network Parameters | 24 | N/A | N/A | 33.4±27.7 | 422.4±144.5 | **6** |
| | Execution Time | 0.21±0.16 | 4.19 | 2.2 | 0.59±0.5 | 0.7±0.4 | **0.16±0.008** |
| | Ensemble Size | 1 | 5 | 5 | N/A | N/A | **1** |
| SUSY | Classification Rate | **0.77±0.04** | Terminated | | 0.73±0.06 | 0.74±0.06 | 0.76±0.04 |
| | Fuzzy Rule | 3.8±1.5 | | | **1.96±0.26** | 5.2±1.5 | 2.84±1.4 |
| | Input Attribute | **1** | | | 18 | 18 | 1 |
| | Network Parameters | **22.9±9.1** | | | 748±33.3 | 3875.2±1119 | 34.1±17.3 |
| | Execution Time | 0.29±0.08 | | | 0.79±0.3 | 2.42±0.6 | **0.6±0.28** |
| | Ensemble Size | **1.9±0.7** | | | N/A | N/A | 1.6±0.7 |

Consolidated algorithms are assessed in six evaluation criteria, namely classification rate, fuzzy rule, input attribute, network parameters, execution time and ensemble size. Classification rate refers to accuracy on testing samples defined as the rate of correctly classified testing samples while fuzzy rule for pENsemble is inspected from a total number of fuzzy rules across all local experts. Input attributes in pENsemble are sampled dynamically in every training instance by assigning crisp weights where a desired number of input attributes is predetermined before process runs whereas input attributes in other algorithms conversely happens to be fixed. Network parameters are enumerated as a total number of network parameters across all local experts and are determined by the type of network architecture. Structural complexities of the base classifiers have been discussed in [24] and are not recounted here. Execution time is obtained from the running time to accomplish a training process, while the ensemble size is measured from the number of base classifiers deployed in the training process. The simulation follows the periodic hold-out process where data streams are generated chunk by chunk in a number of time stamps and each data chunk are partitioned into two parts, namely training and testing and are processed in online mode. The evolution of classification rate, ensemble size of benchmarked algorithms and fuzzy rule evolution of base classifier are visualized in Fig. 2. The classifier is evaluated in the training-then-testing mode and classifier's accuracy is seen from its generalization aptitude in classifying testing samples. That is, it is assumed that data streams arrive in chunk-by-chunk mode where every chunk is split into two parts: training and testing. The performance of the algorithms are measured based on their overall performance across all data streams. The procedure of the periodic hold-out process is depicted in Fig. 3. Numerical results of consolidated algorithms are tabulated in Table 2 and are averaged over the number of time stamps.

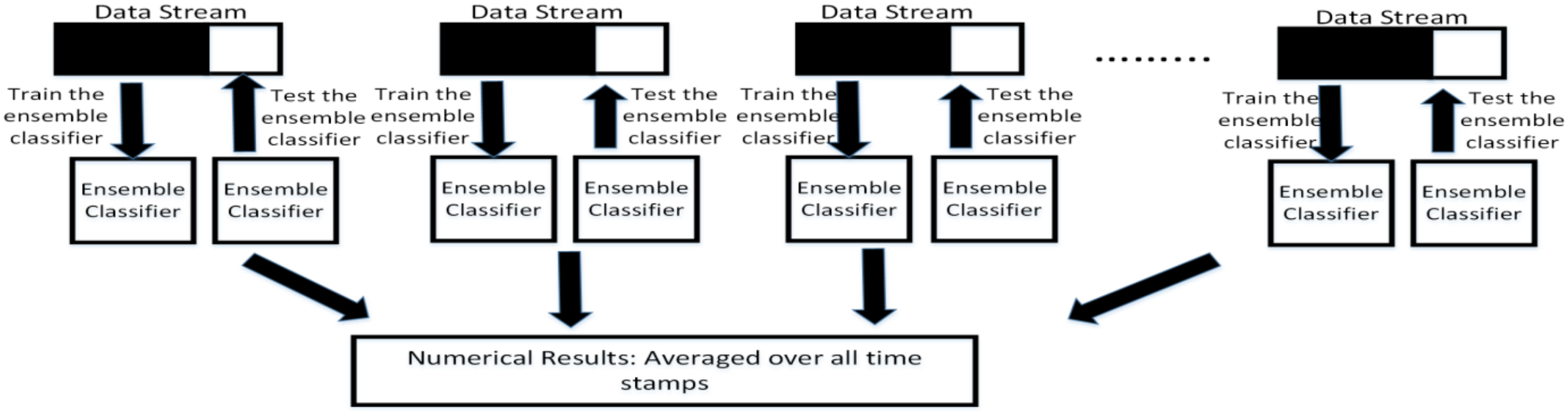

Fig. 3 the principle of periodic hold-out process

From Table 2, pENsemble outperforms its counterparts in the viewpoint accuracy where it produces the highest accuracy in 9 of 15 study cases. It is depicted that pENsemble delivers almost 10% improvement of classification rate compared to its single model version - pClass. pENsemble's accuracy is inferior to Learn++NSE and Learn++CDE in sinH, Gaussian and iris+ data streams. Nonetheless, it is understood that both Learn++NSE and Learn++CDE possess intractable structural complexities since the ensemble size grows exponentially as the number of data streams which might not be a wise option in the real-world data stream environments where the total number of data streams is unpredictable and possibly infinite. This drawback is portrayed in the SUSY problems having 5 million samples. Both algorithms were terminated before obtaining any results because they had run for three days in the same computational resources as other algorithms without returning any results. In the realm of fuzzy rule and network parameters, pENsemble generates a comparable level of complexities even compared to non-ensemble classifiers. These facts are acceptable since pENsemble features two rule pruning scenarios analyzing not only relevance of base classifiers but also approximation of generalization performance of base classifiers. Moreover, the dynamic online feature selection scenario contributes substantially to lower network parameters without compromising the predictive accuracy. The compact and parsimonious structures of pENsemble expedite its execution times which happened to be comparable with its single model counterparts and even faster than them in some study cases. Note that the claim of execution time can be made because all consolidated algorithms were executed under the same computing platform. pENsemble overcomes both Learn++NSE and Learn++CDE in the context of ensemble size in all study cases. It is worth noting that pENsemble makes use of the drift detection method controlling the growth of ensemble structures. The drift detection method brings a step forward from Learn++.NSE and Learn++.CDE since a new data stream does not necessarily trigger the introduction of a new local expert and a new local expert is added only when the conflict attributed to the concept change is severe enough and beyond the scope of existing local experts. This scenario leads to a more resilient approach to deal with the plasticity-stability dilemma than static ensemble or greedy ensemble [27]. Nonetheless, it is recognized that ensemble structure incurs more demanding complexities than a single classifier. Another important observation is in the influence of base-classifier to the overall ensemble performance. It is evident that pENsemble performs similarly under the two versions of pClass in terms of accuracy. Although it is expected that the multivariate Gaussian function imposes high network parameters to be stored in the memory, this type of rule antecedent evolves lower numbers of fuzzy rules than its counterpart – the axis-parallel rule. This leads to comparable memory demand and computational cost.

Fig 2(a) displays evolution of classification rate of benchmarked algorithms in iris+ problem. The iris+ problem has been modified from its standard version by duplicating original samples and randomly selecting samples. It is demonstrated that predictive accuracy of benchmarked algorithms suffers from concept drift and learning strategies help to recover from significant performance deterioration. Fig 2(b) captures the fact that pENsemble has a fully evolvable ensemble structure fully controlled by the drift detection mechanism and ensemble pruning scenario. Compared to Learn++NSE and Learn++CDE, pENsemble incurs substantially more parsimonious ensemble structure since its structure only grows on demand with respect to whether or not drift is present in the data streams and is contrast to Learn++NSE and Learn++CDE where a new base classifier is introduced when a new data stream arrives regardless of whether or not it contains added value to the training process. Fig. 2(c) depicts the evolving nature of base classifier in SEA problem where a base classifier can grow and shrink its structure flexible on the fly. This trait is fruitful to deal with the dynamic and evolving nature of base classifiers. pENsemble is constructed by a collection of axis-parallel pClass as the base classifier. Fuzzy rule extracted by this base classifier in sin problem is illustrated:

***IF*** $X_1$ is close to $A_1(-46.7,742.89)$ AND $X_2$ is close to $A_1(3.54,2.68)$ THEN $y_1 = -0.25 - 0.08X_1 + 1.21X_2$, $y_2 = 1.25 + 0.08X_1 - 1.21X_2$

where $A_1(), A_2()$ stand for the Gaussian membership function corresponding to $X_1, X_2$ coordinates and can be associated by a specific linguistic label. $y_1, y_2$ denote the rule consequents associated to two class labels of sin problem. On the other hand, fuzzy rule generated by pClass with the non-axis parallel Gaussian rule in the sin problem is shown as follows:

***IF*** $X$ is close to $A(\begin{bmatrix} 0.21 \\ 0.35 \end{bmatrix}, \begin{bmatrix} 10,10 \\ 10,5.4 \end{bmatrix})$ THEN

$$y_1 = -0.226 - 0.086X_1 + 1.181X_2,$$

$$y_2 = 1.223 + 0.086X_1 - 1.178X_2$$

where $A()$ refers to the rule antecedent of multivariate Gaussian function which comprises two elements, the centroid and the inverse covariance matrix. It is evident that pClass is constructed under two base-classifiers with different types of fuzzy rules. The multivariate Gaussian rule carries some advantage in terms of its aptitude in dealing with complex data distribution but it comes with the cost of rule transparency due to the absence of atomic clauses of the fuzzy rules. Furthermore, the transformation strategy of multivariate Gaussian rule has to be carried out when executing the ensemble pruning strategy (Section C.2).

## VI. Conclusion

This paper presents a novel evolving ensemble classifier, termed parsimonious ensemble (pENsemble). pENsemble feature some unique characteristics where an evolving classifier, namely pClass, is utilized as its local expert. The flexible working principle of pClass helps pENsemble to handle local drift of data streams effectively because it features an open structure and a fully online working principle. pENsemble constitutes a fully evolving ensemble classifier where its structure is automatically generated and self-expands when a concept drift is detected. pENsemble offers a parsimonious working principle which is resulted from pruning activities of inactive classifiers. It is equipped with two ensemble pruning strategies which assess relevance and generalization power of a local expert. An online feature selection strategy is incorporated into pENsemble. This mechanism actively selects a subset of input attributes and differs from common practice in the literature because it allows to arrive at different subsets of input attributes in every training observation. The efficacy of pENsemble has been numerically validated through 15 real-world and synthetic data streams. It has been compared with 4 well-known algorithms where our algorithm delivers the highest accuracy in 9 of 15 study cases. It is also found that pENsemble generated comparable complexities from those of single classifier variants and far less complexities than those of ensemble classifier variants. Simulations with two different local experts have been carried out to examine its sensitivity to the local expert and pENsemble performs similarly under the two local experts. Future work will be directed toward investigation of granular computing for data stream analytics to address high-level data stream abstraction.

## Acknowledgement

Authors thank Prof. Plamen Angelov for thorough discussion about history of EIS. Authors also thank Dr. Agus Salim for sharing his datasets.

## References

[1] J. Gama, Knowledge Discovery from Data Streams, Chapman & Hall/CRC, Boca Raton, Florida, 2010

[2] P. Angelov, " Autonomous Learning Systems: From Data Streams to Knowledge in Real-time", John Wiley and Sons Ltd., 2012

[3] M. Sayed-Mouchaweh and E. Lughofer, Learning in Non-Stationary Environments: Methods and Applications, Springer, New York, 2012

[4] M. Pratama, J. Lu, E. Lughofer, G. Zhang and M.J. Er, Incremental Learning of Concept Drift Using Evolving Type-2 Recurrent Fuzzy Neural Network, IEEE Transactions on Fuzzy Systems, on-line and in press, 2017

[5] G. Ditzler, et al, " Learning in Nonstationary Environments: A Survey", IEEE Computational Intelligence Magazine, Vol.10(4), pp. 12-25, (2015)

[6] R. M. French, Catastrophic forgetting in connectionist networks, Trends in Cognitive Sciences, vol. 3 (4), pp. 128--135, 1999

[7] P.Angelov and D. Filev, "An approach to online identification of Takagi-Sugeno fuzzy models," IEEE Transactions on Systems, Man, and Cybernetics, Part B. vol. 34, pp. 484-498. 2004

[8] S.W.Tung, C.Quek, C.Guan, "eT2FIS: An Evolving Type-2 Neural Fuzzy Inference System", Information Sciences, vol.220, pp.124-148, (2013)

[9] N. Kasabov, and Q. Song, DENFIS: dynamic evolving neural-fuzzy inference system and its application for time series prediction, IEEE Transactions on Fuzzy Systems .vol10 (2).pp. 144–154. (2002)

[10] M. Pratama, J. Lu, G.Zhang, " Evolving Type-2 Fuzzy Classifier", online and in press, IEEE Transactions on Fuzzy Systems, on line and in press, (2015)

[11] A. Lemos, et al, Adaptive fault detection and diagnosis using an evolving fuzzy classifier, Information Sciences, vol. 220, pp. 64-85, (2013)

[12] P. Brazdil, C. Giraud-Carrier, C. Soares and R. Vilalta, Metalearning, Springer, Berlin Heidelberg, 2009

[13] L. Rokach, Ensemble-based classifiers. Artificial Intelligence Review, vol. 33 (1-2), pp. *1–39*, 2010

[14] J.A Iglesias, A. Ledezma, A. Sanchiz, "Ensemble Method Based on Individual Evolving Classifiers", in 2013 Evolving and Adaptive Intelligent Systems, pp. 78-83, 2013

[15] J.A Iglesias, A. Ledezma, A. Sanchiz, "Analyzing the structure of ensembles based-on evolving classifiers", in 2013 FINO/CAEPIA, 2013

[16] A. Bouchachia, et al, DELA: A Dynamic Online Ensemble Learning Algortihm, in European Symposium on Artificial Neural Networks, Computational Intelligence and Machine Learning, pp. 491- 496, 2014

[17] L. Kuncheva, " Classifiers Ensemble for Changing Environments", Lecture Notes on Computer Sciences, Vol. 3077, pp. 1-15, 2004

[18] R. Elwell and R. Polikar. Incremental learning of concept drift in nonstationary environments. IEEE Transactions on Neural Networks, Vol. 22(10), pp. 1517–1531, 2011

[19] B. Mirza, Z. Lin, and N. Liu, "Ensemble of subset online sequential extreme learning machine for class imbalance and concept drift," Neurocomputing, vol. 149, pp. 315–329, 2015

[20] A. Shaker et al, Self-Adaptive and Local Strategies for a Smooth Treatment of Drifts in Data Streams, Evolving Systems, vol. 5 (4), pp. 239--257, 2014

[21] M. Pratama, J. Lu, E. Lughofer, G. Zhang and S. Anavatti, Scaffolding Type-2 Classifier for Incremental Learning under Concept Drifts, Neurocomputing, vol. 191, pp. 304--329, 2016

[22] P.P.K. Chan, X. Zeng, E. C. C. Tsang, D. S. Yeung, J. W. T. Lee, " Neural Network Ensemble Pruning Using Sensitivity Measure in Web Applications", in IEEE International Conference on Systems, Man and Cybernetics, pp. 3051-3056, 2007

[23] E. Lughofer, P. Angelov," Handling Drifts and Shifts in On-Line Data Streams with Evolving Fuzzy Systems", Applied Soft Computing, vol. 11(2), pp. 2057-2068, 2011

[24] M. Pratama, S.G. Anavatti, M.J. Er and E. Lughofer, pClass: An Effective Classifier for Streaming Examples, IEEE Transactions on Fuzzy Systems, vol. 23 (2), pp. 369--386, 2015

[25] H. Toubakh, M. Sayed-Mouchaweh, " Hybrid dynamic data-driven approach for drift-like fault detection in wind turbines", Evolving Systems, Vol. 6(2), pp. 115-129, 2015

[26] G. Dirzler, R. Polikar, " Hellinger Distance based Drift Detection for Nonstationary Environments", in IEEE Symposium on Computational Intelligence in Computational Intelligence in Dynamic and Uncertain Environments, pp. 41-48, 2011

[27] I. Frias-Blanco, J. D. Campo-Avilla, G. Ramos-Jimenes, R. Morales-Bueno, A. Ortiz-Diaz, Y. Caballero-Mota, "Online and Non-Parametric Drift Detection Methods Based on Hoeffding's Bounds", IEEE Transactions on Knowledge and Data Engineering, Vol. 27(3), pp. 810-823, 2015

[28] D. S. Yeung, W.W. Y. Ng, D. Wang, E. C. C. Tsang, X-Z. Wang, " Localized Generalization Error Model and Its Application to Architecture Selection for Radial Basis Function Neural Network", IEEE Transactions on Neural Networks, Vol. 18(5), pp. 1294-1305, 2007

[29] P. P. Chang, D. S. Yeung, W. W. Y. Ng, C. M. Lin, J. N. K. Liu, " Dynamic Fusion Method Using Localized Generalization Error Model", Information Sciences, Vol. 217, pp. 1-20, 2012
[30] P. P. K. Chan, et al, " Sensitivity Growing and Pruning Method for RBF Networks in Online Learning Environments", in International Conference on Machine Learning and Cybernetics, pp. 1107-1112, 2011
[31] W. W. Y. Ng, A. P. F. Chan, D. S. Yeung, E. C. C. Tsang, " Quantitative Study on the Generalization Error of Multiple Classifier Systems", in IEEE International Conference on Systems, Man and Cybernetics, 2005
[32] J. Wang, P. Zhao, S. Hoi, R. Jin, " Online Feature Selection and Its Applications", IEEE Transactions on Knowledge and Data Engineering, Vol. 26(3), pp. 698-710, 2014
[33] E.Lughofer,"On-line incremental feature weighting in evolving fuzzy classifiers," Fuzzy Sets and Systems, vol. 163(1), pp. 1–23, (2011)
[34] J. Kolter and M. Maloof. Dynamic weighted majority: An ensemble method for drifting concepts. Journal of Machine Learning Research, Vol. 8, pp. 2755–2790, 2007
[35] C. Juang, C. Lin, An on-line self-constructing neural fuzzy inference network and its applications. IEEE Transactions on Fuzzy Systems, vol. 6(1), pp. 12–32, 1998
[36] M.Pratama, S.Anavatti, J.Lu, Recurrent Classifier Based on an Incremental Meta-cognitive-based Scaffolding Algorithm, IEEE Transactions on Fuzzy Systems, Vol.23(6), pp. 2048-2066, 2015
[37] P. Angelov, R. Buswell, " Identification of Evolving Fuzzy Rule-Based Models", IEEE Transactions on Fuzzy Systems, Vol. 10(5), pp. 667-677, 2002
[38] H.-J. Rong, N. Sundarajan, G.-B. Huang, and G.-S. Zhao, "Extended sequential adaptive fuzzy inference system for classification problems," Evolving Systems, vol. 2(2), pp. 71–82, 2011
[39] E. Lughofer, Evolving Fuzzy Systems --- Methodologies, Advanced Concepts and Applications, Springer, Berlin Heidelberg, 2011
[40] M. Pratama, G. Zhang, M-J. Er, S. Anavatti, An Incremental Type-2 Meta-cognitive Extreme Learning Machine, IEEE Transactions on Cybernetics, online and in press, 2016
[41] E. Lughofer, Extensions of Vector Quantization for Incremental Clustering, Pattern Recognition, vol. 41 (3), pp. 995--1011, 2008
[42] E. Lughofer, FLEXFIS: A Robust Incremental Learning Approach for Evolving TS Fuzzy Models, IEEE Transactions on Fuzzy Systems, vol. 16 (6), pp. 1393--1410, 2008
[43] E. Lughofer, et al, On-line Quality Control with Flexible Evolving Fuzzy Systems, in: Learning in Non-Stationary Environments: Methods and Applications, Springer, pp. 375--406, New York, 2012
[44] E. Lughofer, et al, Generalized Smart Evolving Fuzzy Systems, Evolving Systems, Vol. 6 (4), pp. 269--292, 2015
[45] A. Lemos, W. Caminhas and F. Gomide, Multivariable Gaussian Evolving Fuzzy Modeling System, IEEE Transactions on Fuzzy Systems, vol. 19 (1), pp. 91--104, 2011
[46] D. Dovzan, V. Logar and I. Skrjanc, Implementation of an Evolving Fuzzy Model (eFuMo) in a Monitoring System for a Waste-Water Treatment Process, IEEE Transactions on Fuzzy Systems, vol. 23 (5), pp. 1761--1776, 2015
[47] M. Pratama, S.G. Anavatti, P. Angelov and E. Lughofer, PANFIS: A Novel Incremental Learning Machine, IEEE Transactions on Neural Networks and Learning Systems, vol. 25 (1), pp. 55--68, 2014
[48] G.-B. Huang, P. Saratchandran, and N. Sundararajan, "A generalized growing and pruning RBF (GGAP-RBF) neural network for function approximation," IEEE Transactions on Neural Networks, vol. 16(1), pp. 57–67, 2005
[49] H. J. Rong, N. Sundararajan, G. B. Huang, and P. Saratchandran, "Sequential adaptive fuzzy inference system (SAFIS) for nonlinear system identification and time series prediction," Fuzzy Sets and Systems, vol. 157(9), pp. 1260–1275, 2006
[50] J. A. Iglesias, A. Ledezma, A. Sanchis, "An ensemble method based on evolving classifiers: eStacking ", in IEEE Symposium on Evolving and Autonomous Learning System, pp. 124-131, 2014
[51] E. Lughofer et al, Reliable All-Pairs Evolving Fuzzy Classifiers, IEEE Transactions on Fuzzy Systems, vol. 21 (4), pp. 625--641, 2013
[52] G. Ditzler and R. Polikar,"Incremental learning of concept drift from streaming imbalanced data," in IEEE Transactions on Knowledge & Data Engineering, vol. 25(10), pp. 2283–2301, 2013
[53] J. Gama, P. Medas, G. Castillo, and P. Rodrigues, "Learning with drift detection," in Proceeding of Brazilian Symposium on Artificial Intelligence., vol. 3171, pp. 286–295, 2004
[54] K. S. Yap, et al, "Improved GART neural network model for pattern classification and rule extraction with application to power system," IEEE Transactions on Neural Networks, vol. 22(12), pp. 2310–2323, 2011
[55] B. Vigdor and B. Lerner, "The Bayesian ARTMAP," IEEE Transactions Neural Networks, vol. 18(6), pp. 1628–1644, 2007
[56] J.-C. de Barros and A. L. Dexter, "On-line identification of computationally undemanding evolving fuzzy models," Fuzzy Sets and Systems, vol. 158, pp. 1997–2012, 2007
[57] Y. Xu, K. W. Wong, and C. S. Leung, "Generalized recursive least square to the training of neural network," IEEE Transactions on Neural Networks, vol. 17(1), pp. 19–34, 2006
[58] R. Polikar, L. Udpa, S. Udpa, V. Honavar, "Learn++: An incremental learning algorithm for supervised neural networks," IEEE Transactions on System, Man and Cybernetics (C), Special Issue on Knowledge Management, vol. 31(4), pp. 497-508, 2001
[59] L. L. Minku and X. Yao, "DDD: A new ensemble approach for dealing with drifts," IEEE Transactions on Knowledge and Data Engineering, vol. 24(4), pp. 619–633, 2012
[60] L. L. Minku, A. P. White, and X. Yao, "The impact of diversity on online ensemble learning in the presence concept of drift," IEEE Transactions on Knowledge and Data Engineering, vol. 22(5), pp. 730–742, 2010
[61] W.N. Street and Y. Kim, "A Streaming Ensemble Algorithm (SEA) for Large-Scale Classification," in International Conference on Knowledge Discovery and Data Mining, pp. 377-382, 2001
[62] K. Subramanian, S. Suresh, N. Sundararajan, "A metacognitive neuro-fuzzy inference system (McFIS) for sequential classification problems", IEEE Transactions on Fuzzy Systems, Vol. 21(6), pp. 1080-1095, 2013
[63] Agus Salim, et al, " C-reactive protein and serum creatinine, but not haemoglobin A1c, are independent predictors of coronary heart disease risk in non-diabetic Chinese", European journal of preventive cardiology, Vol. 23(12), pp. 1339-1349, 2016
[64] D. E. Sr. Dimla, P.M. Lister, " On-line Metal Cutting Tool Condition Monitoring. II: Tool-state Classification using Multi-Layer Perceptron Neural Networks", International Journal of Machine Tools and Manufacture, Vol. 40, 769-781, 2000
[65] P. Angelov, et al, "Evolving fuzzy classifiers using different model architectures", Fuzzy Sets and Systems, Vol.159(23) ,pp.3160– 3182, 2008
[66] P.Angelov et al, "Evolving fuzzy rule-based classifiers from data streams," IEEE Transactions on Fuzzy Systems, vol. 16(6), pp. 1462–1475, 2008
[67] R. D. Baruah, P. Angelov, J. Andreu, "Simpl_eClass: Simplified potential-free evolving fuzzy rule-based classifiers," in Proceeding of IEEE International Conference on Systems, Man and Cybernetics, Anchorage, AK, USA, Oct. 7–9, 2011, pp. 2249–2254
[68] D. Kangin, P. Angelov, J. A. Iglesias, " Autonomously Evolving Classifier TEDAClass", Information Sciences, Vol. 366, pp. 1-11, 2016
[69] D. Kangin, P. Angelov, J. A. Iglesias, A. Sanchis, " Evolving Classifier TEDAClass for Big Data", in Proceeding of INNS conference on Big Data, pp. 9-18, 2015
[70] P. Angelov, X. Gu, " MICE: Multi-layer multi-model images classifier ensemble", In proceeding of IEEE International Conference on Cybernetics, 2017
[71] P. Angelov, N. Kasabov, "Evolving Intelligent Systems, eIS", IEEE SMC eNewsletter, Vol.15, pp. 1-13, 2006
[72] M.Pratama, S.Anavatti, E.Lughofer, GENEFIS: Towards an Effective Localist Network, IEEE Transactions on Fuzzy Systems, Vol.2(3), pp.547-562, (2014)
[73] Z. Xie, Y. Xu, Q. Hu, P. Zhu, "Margin distribution based bagging pruning", Neurocomputing, Vol. 85, pp. 11-10, 2012
[74] L. Li, Q. Hu, X. Qu, D. Yu, "Exploration of classification confidence in ensemble learning", Pattern Recognition, Vol. 47, pp. 3120-3130, 2014
[75] G. D. C. Cavalcanti, L. S. Oliviera, T. J. M. Moura, G. V. Calvarho, " Combining diversity measures for ensemble pruning", Pattern Recognition Letters, Vol. 74, pp. 38-45, 2016
[76] Z. Lu, X. Wu, X. Zhu, J. Bongard, " Ensemble Pruning via Individual Contribution Ordering", Proceedings of the 16th ACM SIGKDD international conference on Knowledge discovery and data mining, pp. 871-880, 2010
[77] H. Chen, P. Tino, X. Yao, "Predictive Ensemble Pruning by Expectation Propagation", IEEE Transactions on Knowledge and Data Engineering, Vol. 21(7), pp. 999-1013, 2009
[78] S. Tong, Y. Li, P. Shi, "Observer-Based Adaptive Fuzzy Backstepping Output Feedback Control of Uncertain MIMO Pure-Feedback Nonlinear Systems", IEEE Transactions on Fuzzy Systems, Vol. 20(4), pp. 771 – 785, 2012
[79] Y. Li, S. Tong, Y. Liu, T. Li, "Adaptive Fuzzy Robust Output Feedback Control of Nonlinear Systems With Unknown Dead Zones Based on a Small-Gain Approach", IEEE Transactions on Fuzzy Systems, Vol. 22(1), pp. 164-176, 2014